%% file: egbib.tex
\documentclass[conference]{IEEEtran}
\usepackage{xcolor}         
\usepackage{graphicx}
\usepackage{bm}
\usepackage{makecell}
\usepackage{bbm}
\usepackage{xspace}
\usepackage{float}
\usepackage{color}
\usepackage{colortbl}
\usepackage{dsfont}
\usepackage{multirow}
\usepackage{multicol}
\usepackage{pgfplots}
\usepackage{algorithm}
\usepackage{algpseudocode}
\usepackage{adjustbox}
\usepackage{listings}
\usepackage{bbding}
\usepackage{pifont}
\usepackage{wasysym}
\usepackage{amssymb}

\ifCLASSINFOpdf

\def\eg{\emph{e.g., }}

\def\ie{\emph{i.e., }}

\newlength\savedwidth
\newcommand\whline{\noalign{\global\savedwidth\arrayrulewidth
                           \global\arrayrulewidth 2pt}%
                  \hline
                  \noalign{\global\arrayrulewidth\savedwidth}}
\newlength\savewidth
\newcommand\shline{\noalign{\global\savewidth\arrayrulewidth
                           \global\arrayrulewidth 0.5pt}%
                  \hline
                  \noalign{\global\arrayrulewidth\savewidth}}

\begin{document}
%
\title{ICDAR 2023 Video Text Reading Competition for Dense and Small Text}

\author{\IEEEauthorblockN{Weijia Wu$^{1}$$^{3}$$^{4}$, Yuzhong Zhao$^2$, Zhuang Li$^3$, Jiahong Li$^3$, Mike Zheng Shou$^4$, \\
Umapada Pal$^5$, Dimosthenis Karatzas$^6$, Xiang Bai$^7$}
\IEEEauthorblockA{$^1$Zhejiang University, China; Email: weijiawu@zju.edu.cn\\
$^2$University of Chinese Academy of Sciences, China; Email: zhaoyuzhong20@mails.ucas.ac.cn\\
$^3$Kuaishou Technology, China; Email: \{lizhuang05,lijiahong\}@kuaishou.com\\
$^4$National University of Singapore, Singapore; Email:
weijiawu@u.nus.edu,mikeshou@nus.edu.sg\\
$^5$Computer Vision and Pattern Recognition Unit, Indian Statistical Institute, India; Email: umapada@isical.ac.in \\
$^6$Computer Vision Centre, Universitat Autónoma de Barcelona; Email: dimos@cvc.uab.es \\
$^7$Huazhong University of Science and Technology, China; Email: xbai@hust.edu.cn\\
}}


%


\maketitle

\begin{abstract}
Recently, video text detection, tracking and recognition in natural scenes are becoming very popular in the computer vision community. 
However, most existing algorithms and benchmarks focus on common text cases~(\eg normal size, density) and single scenario, while ignore extreme video texts challenges, \ie{} dense and small text in various scenarios.
In this competition report, we establish a video text reading benchmark, named DSText, which focuses on dense and small text reading challenge in the video with various scenarios.
Compared with the previous datasets, the proposed dataset mainly include three new challenges: 
1) Dense video texts, new challenge for video text spotter.
2) High-proportioned small texts.
3) Various new scenarios, \eg{} `Game', `Sports', etc. 
The proposed DSText includes 100 video clips from 12 open scenarios, supporting two tasks~(\ie{} video text tracking (Task 1) and end-to-end video text spotting (Task2)).
During the competition period (opened on 15th February, 2023 and closed on 20th March, 2023), a total of 24 teams participated in the three proposed tasks with around 30 valid submissions,
respectively.
In this article, we describe detailed statistical information of the dataset, tasks, evaluation protocols and the results summaries of the ICDAR 2023 on DSText competition.
Moreover, we hope the benchmark will promise the video text research in the community.
\end{abstract}


%
\IEEEpeerreviewmaketitle

\section{Introduction}

Video text spotting~\cite{yin2016text} has received increasing attention due to its numerous applications in computer vision, \eg{video understanding~\cite{srivastava2015unsupervised}, video retrieval~\cite{dong2021dual}, video text translation, and license plate recognition~\cite{anagnostopoulos2008license}, etc.}
There already exist some video text spotting benchmarks, which focus on easy cases, \eg{} normal text size, density in single scenario.
ICDAR2015~(Text in Videos)~\cite{karatzas2015icdar}, as the most popular benchmark, was introduced during the ICDAR Robust Reading Competition in 2015 focus on wild scenarios: walking outdoors, searching for a shop in a shopping street, etc. 
YouTube Video Text~(YVT)~\cite{nguyen2014video} contains 30 videos from YouTube. The text category mainly includes overlay text~(caption) and scene text (\eg{driving signs, business signs}).
RoadText-1K~\cite{reddy2020roadtext} provide 1,000 driving videos, which promote driver assistance and self-driving systems.
LSVTD~\cite{cheng2019you} proposes 100 text videos, 13 indoor (\eg{ bookstore, shopping mall}) and 9 outdoor (\eg{highway, city road}) scenarios, and support two languages, \ie{} English and Chinese.
BOVText~\cite{wu2021bilingual} establishes a a large-scale, bilingual video text benchmark, including abundant text types, \ie{} title, caption or scene text.

\begin{figure}[t]
\begin{center}:
\includegraphics[width=0.48\textwidth]{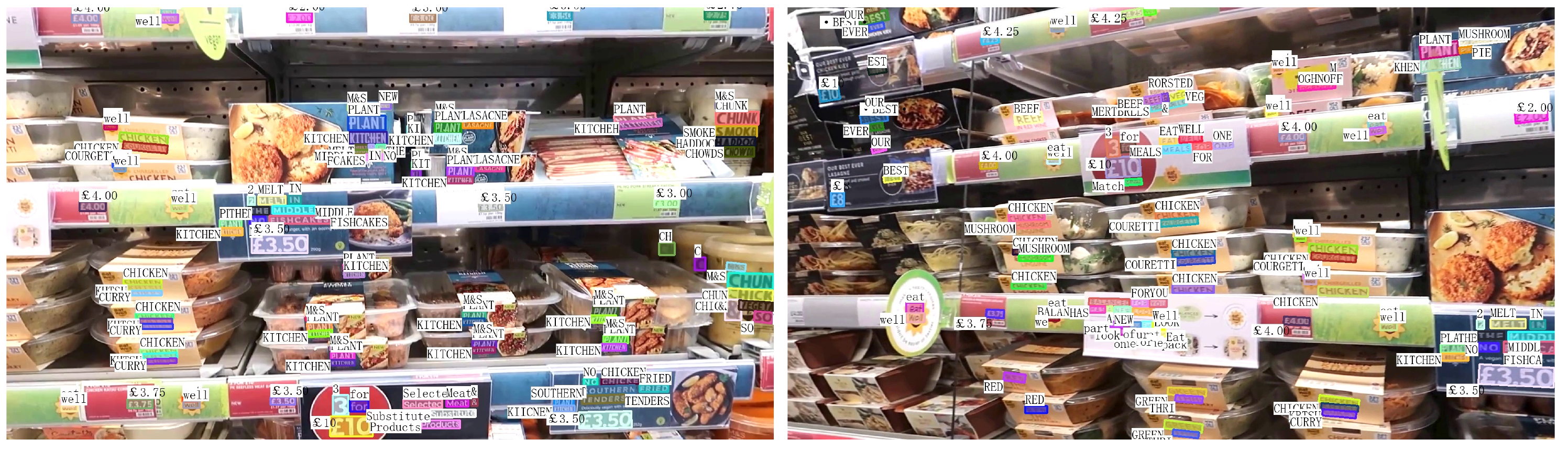}
\vspace{-5mm} 
\caption{\textbf{Visualization of DSText.} Different from previous benchmarks, DSText focuses on dense and small text challenge.}
\label{fig1_vis}
\end{center}
\vspace{-3mm} 
\end{figure}

However, the above benchmarks still suffer from some limitations: 1) Most text instances present normal text size without challenge, \eg{} ICDAR2015(video) YVT, BOVText.
2) Sparse text density in single scenario, \eg{} RoadText-1k and YVT, which can not evaluate the small and dense text robustness of algorithm effectively.
3) Except for ICDAR2015(video), most benchmarks present unsatisfactory maintenance. YVT, RoadText-1k and BOVText all do not launch a corresponding competition and release open-source evaluation script.
Besides, the download links of YVT even have become invalid.
The poor maintenance plan is not helpful to the development of video text tasks in the community.
To break these limitations, we establish one new benchmark, which focus on dense and small texts in various scenarios, as shown in Fig.~\ref{fig1_vis}.
The benchmark mainly supports two tasks, \ie{} \textit{video text tracking}, and end to end \textit{video text spotting} tasks, includes 100 videos with 56k frames and 671k text instances.
%

\begin{table*}[t]
    \centering
	\caption{\textbf{Statistical Comparison.} `Box Type‘ , `Text Area' denote detection box annotation type and average area of text while the shorter side of image is 720 pixels. `Text Density' refers to the average text number per frame. The proposed DSText presents more small and dense texts.}
	\label{table1}
	\input{table/table1.tex}

\end{table*}

Therefore, we organize the ICDAR 2023 Video Text Reading competitive for dense and small text, which generates a large-scale video text database, and proposes video text tracking, spotting tasks, and corresponding evaluation methods. 
This competition can serve as a standard benchmark for assessing the robustness of algorithms that are designed for video text spotting in complex natural scenes,
which is more challenging.
The proposed competition and dataset will enhance the related direction~(Video OCR) of the ICDAR community from two main aspects:
\begin{itemize}
    \item  Compared to the current existing video text reading datasets, the proposed DSText has some special features and challenges, including 1) Abundant scenarios, 2) higher proportion of small text, 3) dense text distribution. Tab.~\ref{table1},  Fig.~\ref{fig4_vis}, Fig.~\ref{fig5_vis}, and Fig.~\ref{fig5_vis1} present detailed statistical comparison and analysis.
    \item  The competition supports two tasks: video text tracking and end-to-end video text spotting. And we provide comprehensive evaluation metrics, including $\rm ID_P$, $\rm ID_R$, $\rm ID_{F1}$~\cite{dendorfer2019cvpr19}, MOTA, and MOTP. These metrics are widely used on previous video text benchmarks, such as ICDAR2015~\cite{karatzas2013icdar,zhou2015icdar}. 
    We are proud to report the successful completion of the competition, which has garnered over 25 submissions and attracted wide interest. The submissions have inspired new insights, ideas, and approaches, which promise to advance the state of the art in video text analysis.
\end{itemize}

\section{Competition Organization}
ICDAR 2023 video text reading competition for dense and small text is organized by a joint team, including Zhejiang University, University of Chinese Academy of Sciences, Kuaishou Technology, National University of Singapore, Computer Vision and Pattern Recognition Unit, Computer Vision Centre, Universitat Autónoma de Barcelona, and Huazhong University of Science and Technology.
And we organize the competition on the Robust Reading Competition Website~\footnote{https://rrc.cvc.uab.es/?ch=22\&com=introduction}, where provide corresponding download links of the datasets, and user interfaces for participants and submission page for their results.

\subsection{Competition Schedule}

The official competition schedule is as follows:

\begin{itemize}
\item December 1, 2022: Website online.

\item February 1, 2023: Sample training videos available.

\item February 15, 2023: Release of full training set and ground truth.

\item March 15, 2023: Test set is available, and website opens for results submission.

\item March 20, 2023: Deadline of the competition and result submission closes.

\item March 31, 2023: Submission deadline for 1-page competition report, and the final ranking will be released after checking the results.

\item August 21-26, 2023: Presentation or results at ICDAR 2023 Conference.

\end{itemize} 

Overall, after removing the duplicate submissions, we received 30 valid submissions from 24 teams from both research communities and industries for the three tasks. 


%

\begin{figure}
\begin{center}
\includegraphics[width=0.48\textwidth]{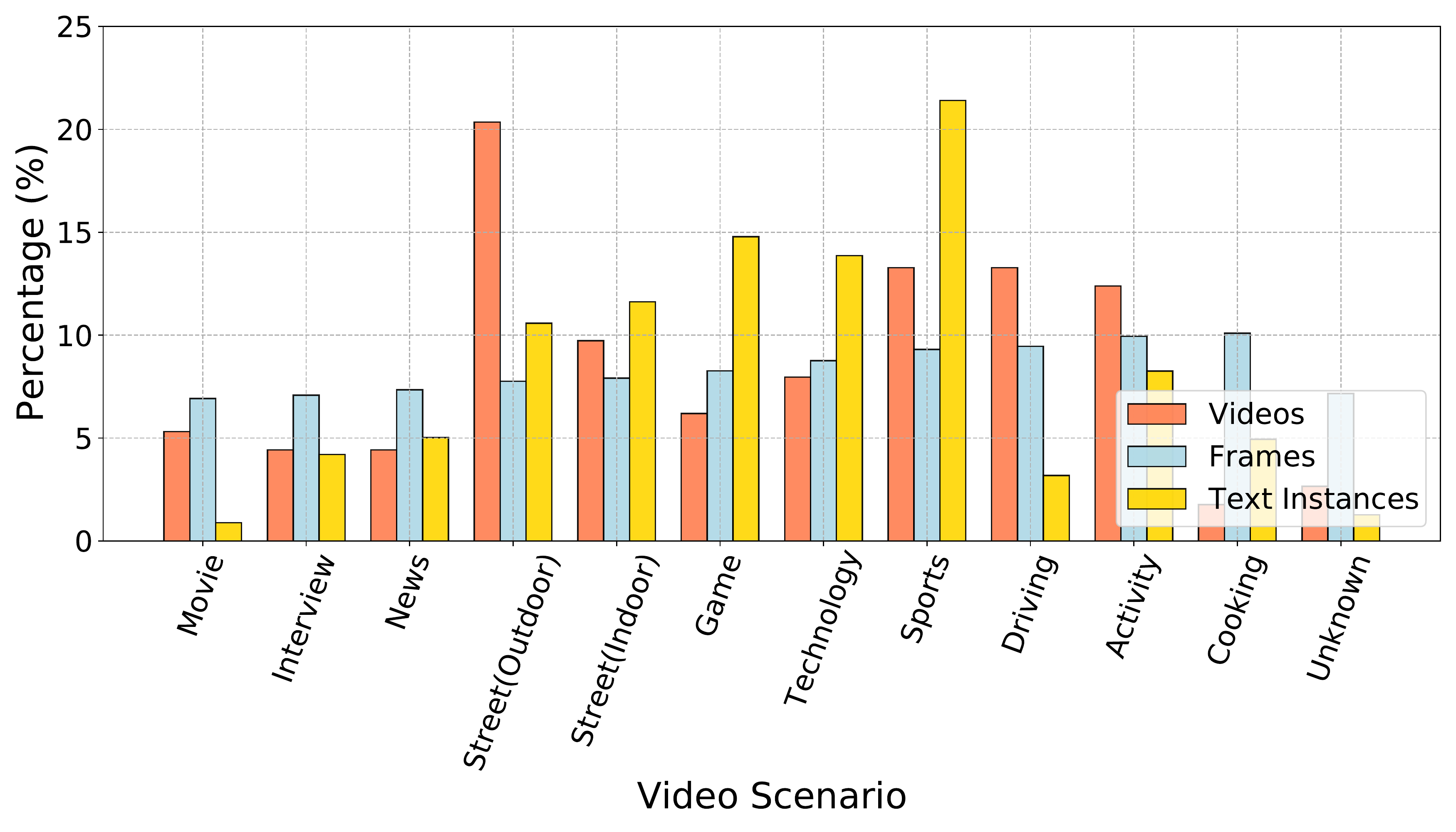}
\caption{\textbf{The Data Distribution for 12 Open Scenarios.} "\%" denotes the percentage of each scenario data over the whole data.}
\label{fig4_vis}
\end{center}
\end{figure}

\section{Dataset}
\subsection{Dataset and Annotations}
\textbf{Dataset Source.} The videos in DSText are collected from three parts: 1) $30$ videos sampled from the large-scale video text dataset BOVText~\cite{wu2021bilingual}.
BOVText, as the largest video text dataset with various scenarios, includes a mass of small and dense text videos.
We select the top $30$ videos with small and dense texts via the average text area of the video and the average number of text per frame.   
2) $10$ videos for driving scenario are collected from RoadText-1k~\cite{reddy2020roadtext}. 
As shown in Fig.~\ref{table1}, RoadText-1k contains abundant small texts, thus we also select $10$ videos to enrich the driving scenario.
3) $60$ videos for street view scenes are collected from YouTube.
Except for BOVText and RoadText-1k, we also obtain $60$ videos with dense and small texts from YouTube, which mainly cover street view scenarios.
Therefore, we obtain $100$ videos with $56$k video frames, as shown in Table~\ref{table1}.
Then the dataset is divided into two parts: the training set with $29,107$ frames from $50$ videos, and the testing set with $27,234$ frames from $50$ videos.

\textbf{Annotation.} For these videos from BOVText, we just adopt the original annotation, which includes four kinds of description information: the rotated bounding box of detection, the tracking identification(ID) of the same text, the content of the text for recognition, the category of text, \ie{} caption, title, scene text, or others.
As for others from RoadText-1k and YouTube, we hire a professional annotation team to label each text for each frame.
The annotation format is the same as BOVText.
One mentionable point is that the videos from RoadText-1k only provides the upright bounding box~(two points), thus we abandon the original annotation and annotate these videos with the oriented bounding box.
Due to the structure of the video source, it is not allowed to use BOVText and RoadText-1k as extra data for training in this competition.
As a \textit{labor-intensive} job, the whole labeling process takes \textbf{30} men in one months, \ie{} around \textbf{4,800} man-hours, to complete the 70 video frame annotations.
As shown in Fig.~\ref{fig1_vis_case}, it is quite time-consuming and expensive to annotate a mass of text instances at each frame.

\begin{figure}
\begin{center}
\includegraphics[width=0.48\textwidth]{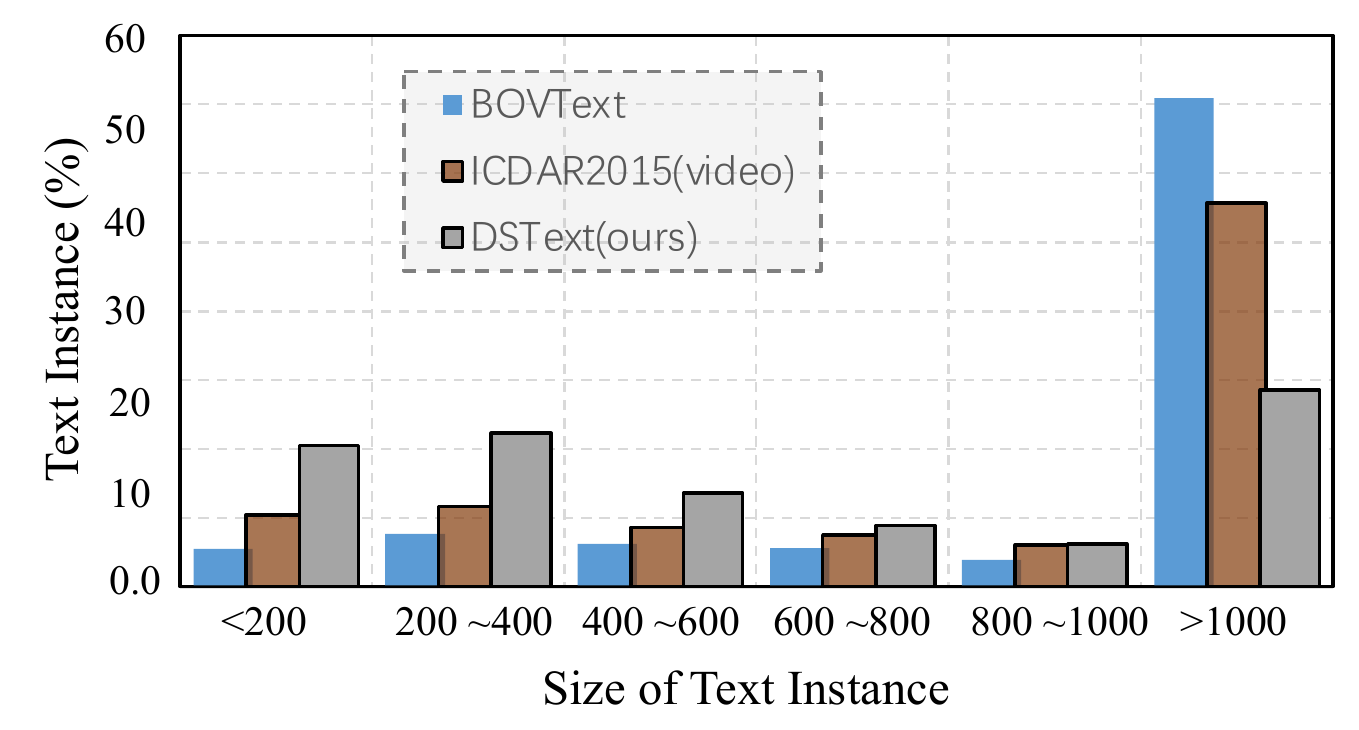}
\caption{\textbf{The distribution of different text size range on different datasets} "\%" denotes the percentage of text size region over the whole data. Text area (\# pixels) is calculated while the shorter side of image is 720 pixels.}
\label{fig5_vis}
\end{center}
\end{figure}

\begin{figure*}[t]
\begin{center}:
\includegraphics[width=1\textwidth]{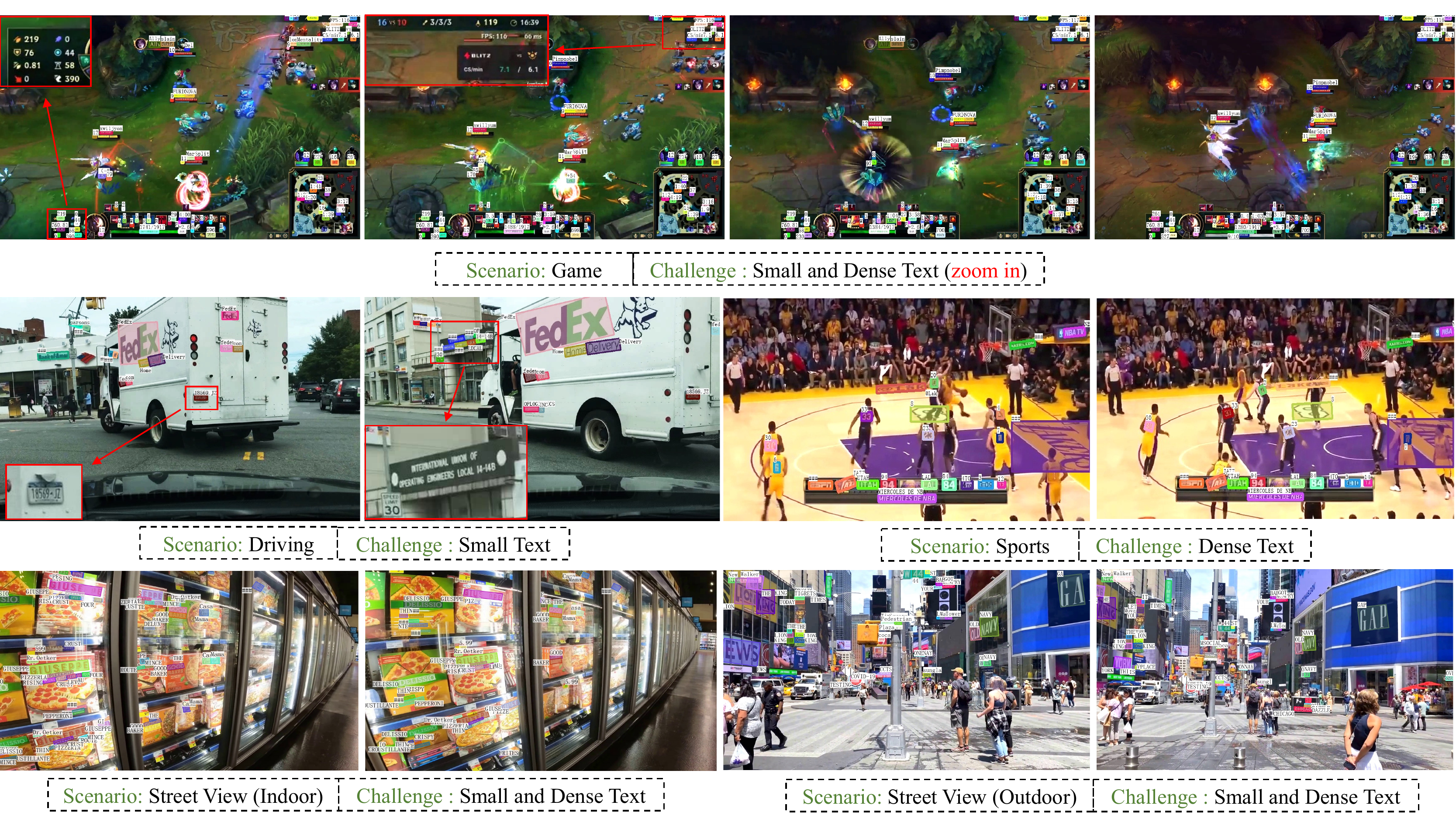}
\vspace{-5mm} 
\caption{\textbf{More Qualitative Video Text Visualization of DSText.} DSText covers small and dense texts in various scenarios, which is more challenge.}
\label{fig3_vis}
\end{center}
\vspace{-3mm} 
\end{figure*}

\subsection{Dataset Comparison and Analysis}

\begin{figure}
\begin{center}
\includegraphics[width=0.48\textwidth]{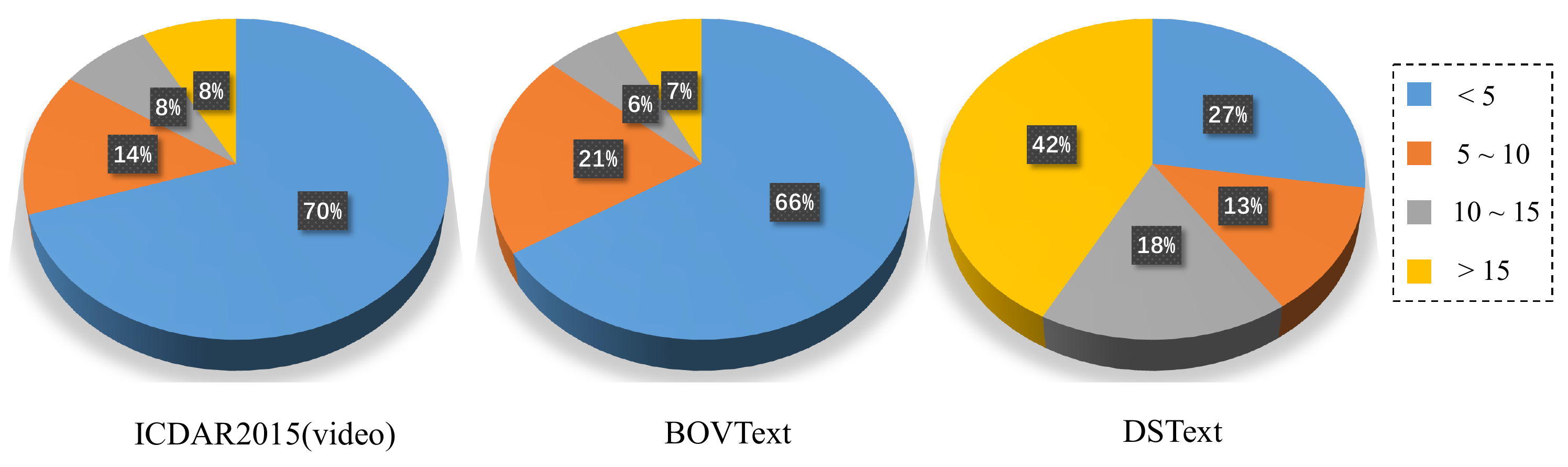}
\caption{\textbf{Comparison for frame percentage of different text numbers.} "\%" denotes the percentage of the corresponding frame over the whole data.}
\label{fig5_vis1}
\end{center}
\end{figure}

The statistical comparison and analysis are presented by three figures and one table.
Table.~\ref{fig4_vis} presents an overall comparison for the basic information, \eg{} number of video, frame, text, and supported scenarios.
In comparison with previous works, the proposed DSText shows the denser text instances density per frame ~(\ie{} average $23.5$ texts per frame) and smaller text size~(\ie{} average $1,984$ pixels area of texts).

\textbf{Video Scenario Attribute}. As shown in Fig.~\ref{fig4_vis}, we present the distribution of video, frame, and frame of $11$ open scenarios and an "Unknown" scenario on DSText.
`Street View~(Outdoor)' and `Sport' scenarios present most video and text numbers, respectively.
And the frame number of each scenario is almost the same.
We also present more visualizations for `Game', `Driving', `Sports' and `Street View' in Fig.~\ref{fig3_vis}.

\textbf{Higher Proportion of Small Text}. Fig.~\ref{fig5_vis} presents the proportion of different text areas. 
The proportion of big text~(more than $1,000$ pixel area) on our DSText is less than that of BOVText and ICDAR2015(video) with at least $20\%$.
Meanwhile, DSText presents a higher proportion for small texts~(less 400 pixels) with around $22\%$. 
As shown in Table.~\ref{table1}, RoadText-1k~\cite{reddy2020roadtext} and LSVTD~\cite{cheng2019you} also show low average text area, but their text density is quite sparse~(only $0.75$ texts and $5.12$ per frame), and RoadText-1k only focuses on the driving domain, which limits the evaluation of other scenarios.

\begin{figure}[t]
\begin{center}:
\includegraphics[width=0.48\textwidth]{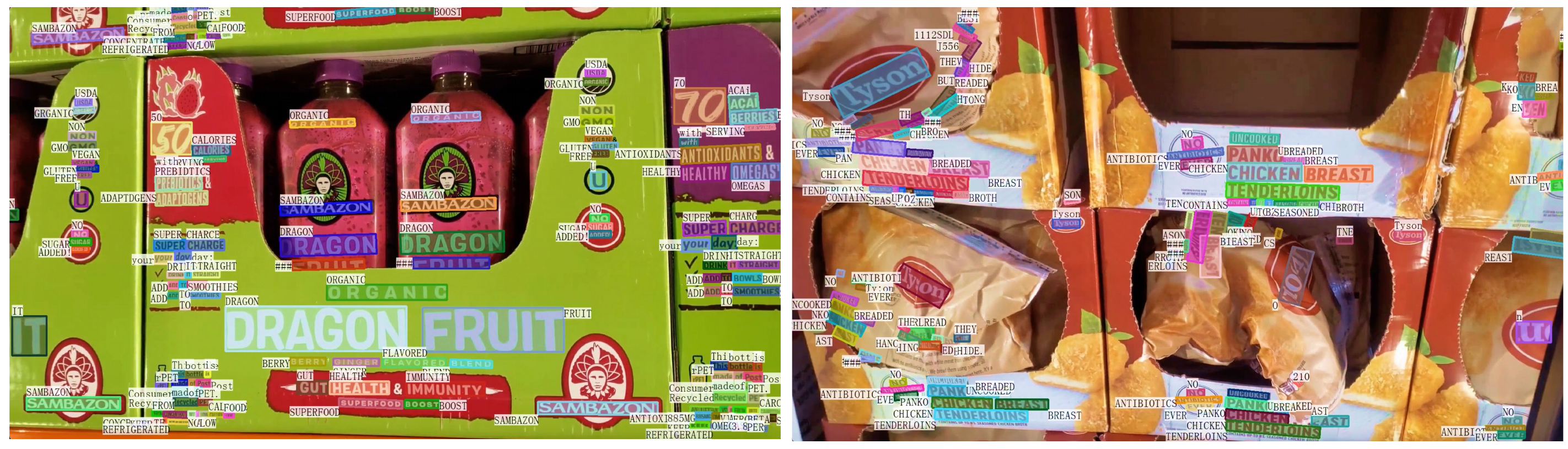}
\vspace{-5mm} 
\caption{\textbf{One Case for around 200 Texts per Frames.} DSText includes huge amounts of small and dense text scenarios, which is a new challenge.}
\label{fig1_vis_case}
\end{center}
\vspace{-3mm} 
\end{figure}

\textbf{Dense Text Distribution}. Fig.~\ref{fig5_vis1} presents the distribution of text density at each frame.
The frame with more than $15$ text instances occupies $42\%$ in our dataset, at least $30\%$ improvement than the previous work, which presents more dense text scenarios.
Besides, the proportion of the frame with less $5$ text instances is just half of the previous benchmarks, \ie{} BOVText, and ICDAR2015(video).
Therefore, the proposed DSText shows the challenge of dense text tracking and recognition. 
More visualization can be found in Fig.~\ref{fig3_vis}~(Visualization for various scenarios) and Fig.~\ref{fig1_vis_case} (Representative case with around 200 texts per frame).

\textbf{WordCloud}. We also visualize the word cloud for text content in Fig~\ref{fig5_vis1}.
All words from annotation must contain at least $3$ characters, we consider the words less four characters usually are insignificant, \eg{} `is'.

\section{Tasks and Evaluation Protocols}
The competition include two tasks: 1) the video text tracking, where the objective is to localize and track all words in the video sequences. and 2) the end-to-end video text spotting: where the objective is to localize, track and recognize all words in the video sequence.

\label{metric}
\textbf{Task 1: Video Text Tracking.}
In this task, all the videos~(50 train videos and 50 test videos) will be provided as MP4 files.
Similar to ICDAR2015 Video~\cite{zhou2015icdar}, the ground truth will be provided as a single XML file per video.
A single compressed (zip or rar) file should be submitted containing all the result files for all the videos of the test set. 

The task requires one network to detect and track text over the video sequence simultaneously.
Given an input video, the network should produce two results: a rotated detection box, and tracking ids of the same text.
For simplicity, we adopt the evaluation method from the ICDAR 2015 Video Robust Reading competition~\cite{zhou2015icdar} for the task.
The evaluation is based on an adaptation of the MOTChallenge~\cite{dendorfer2019cvpr19} for multiple object tracking. 
For each method, MOTChallenge provides three different metrics: the Multiple Object Tracking Precision (MOTP), the Multiple Object Tracking Accuracy (MOTA), and the IDF1. See the 2013 competition report~\cite{karatzas2013icdar} and MOTChallenge~\cite{dendorfer2019cvpr19} for details about these metrics.
%
%
In our competition, we reuse the evaluation scripts from the 2015  video text reading competition~\cite{zhou2015icdar},
and transfer the format of annotation to the same as that of the ICDAR2015 Video.
%

\textbf{Task 2: End-to-End Video Text Spotting.}
Video Text Spotting~(VTS) task that requires simultaneously detecting, tracking, and recognizing text in the video. 
The word recognition performance is evaluated by simply whether a word recognition result is completely correct. And the word recognition evaluation is case-insensitive and accent-insensitive. All non-alphanumeric characters are not taken into account, including decimal points, such as '1.9' will be transferred to ’19‘ in our GT. 
Similarly, the evaluation method~(\ie{} ${\rm ID_{F1}}$, ${\rm MOTA}$ and ${\rm MOTP}$) from the ICDAR 2015 Robust Reading competition is also adopted for the task.
%
%
In the training set, we provide the detection coordinates, tracking id, and transcription result. 

Note: From 2020, the ICDAR 2015 Robust Reading competition online evaluation~\footnote{https://rrc.cvc.uab.es/?ch=3\&com=evaluation\&task=1} has updated the evaluation method, and added one new metric, \ie{} ID metrics~($ID_{F1}$)~\cite{li2009learning,ristani2016performance}.
Similarly, we adopted the updated metric for the two tasks. 

\section{Baseline: TransDETR}
To help participants more easily engage in our competition, we have also provided corresponding baseline algorithms in the competition website~\footnote{https://rrc.cvc.uab.es/?ch=22\&com=downloads}, \ie{} TransDETR~\cite{wu2022end}.
where including the corresponding training and inference code for the competition.
TransDETR is a novel, simple and end to end video text DEtection, Tracking, and Recognition framework~(TransDETR), which views the video text spotting task as a direct long-sequence temporal modeling problem.

\begin{figure}
\begin{center}
\includegraphics[width=0.45\textwidth]{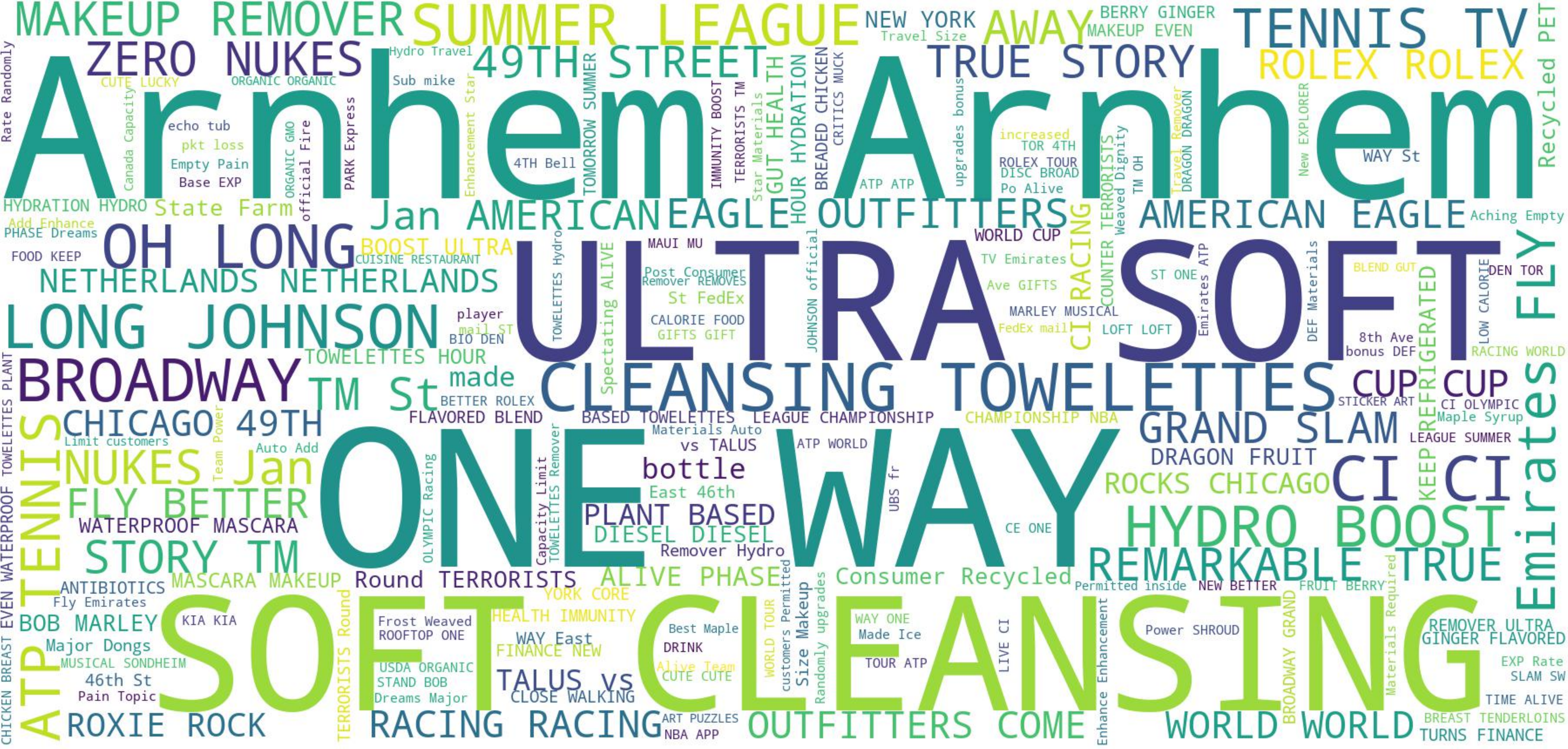}
\caption{\textbf{Wordcloud visualizations for DSText.}}
\label{fig5_vis11}
\end{center}
\end{figure}

\begin{table*}[t]
    \centering
	\caption{\textbf{Task 1: Video Text Tracking Results.} - denotes missing descriptions in affiliations.}
	\label{table2}
	\input{table/task1.tex}
\end{table*}

\begin{table*}[t]
    \centering
	\caption{\textbf{Task2: End-to-End Video Text Spotting Results.} - denotes missing descriptions in affiliations.}
	\label{table3}
	\input{table/task2.tex}
\end{table*}

\section{Submissions}
The result for Task 1 and Task 2 are presented on Table.~\ref{table2} and Table.~\ref{table3}, respectively.

\subsection{Top 3 Submissions in Task 1}
\textbf{Tencent-OCR team.} The top 1 solution method follows the framework of Cascade Mask R-CNN~\cite{cai2018cascade}. Multiple backbones including HRNet~\cite{wang2020deep} and InternImage~\cite{wang2022internimage} are used to enhance the performance.
On the text tracking task, the team designed four different metrics to compare the matching similarity between the current frame detection box and the existing text trajectory, \ie{} box
IoU, text content similarity, box size proximity and text geometric neighborhood relationship
measurement. 
These matching confidence scores are used as a weighted sum for the matching
cost between the currently detected box and tracklet. 
When there is a time difference between
the current detection box and the last appearance of the tracklet, the IoU and box size metrics
are divided by the corresponding frame number difference to prioritize matching with the latest
detection box in the trajectory set. 
They construct a cost matrix for each detected box and existing trajectory in each frame, where the Kuhn-Munkres algorithm is used to obtain matching pairs. 
When the metrics are less than a certain threshold, their corresponding costs are set to 0. 
Finally, they perform grid search to find better hyperparameters. 
Referring to ByteTrack~\cite{zhang2022bytetrack}, boxes with high detection/recognition scores are prioritized for matching, followed by boxes with lower detection/recognition scores. 
Each box that is not linked to an existing trajectory is only considered as a starting point for a new trajectory when its detection/recognition score is high enough. %
Finally, we removed low-quality trajectories with low text confidence scores and noise trajectories with only one detection box. 

Strong data augmentation strategies are adopted such as photometric distortions, random
motion blur, random rotation, random crop, and random horizontal flip. 
And IC13~\cite{karatzas2013icdar}, IC15~\cite{karatzas2015icdar}, IC15 Video~\cite{karatzas2015icdar} and Synth800k~\cite{synthtext} are involved during the training phase. Furthermore, they treat non-alphanumeric characters as negative samples and regard text instances that are labeled `\#\#DONT\#CARE\#\#' as ignored ones during the training phase. In the inference phase, they use multiple resolutions of 600, 800, 1000, 1333, 1666, and 2000. 

\textbf{Guangzhou Shiyuan Technology team.}
The team utilized Mask R-CNN~\cite{he2017mask} and DBNet~\cite{liao2020real} as their base architectures. These were trained separately, and their prediction polygons were fused through non-maximum suppression. For the tracking stage, VideoTextSCM~\cite{gao2021video} was adopted, with Bot-SORT~\cite{aharon2022bot} replacing the tracker in the VideoTextSCM model. Bot-SORT is an enhanced multi-object tracker that leverages MOT bag-of-tricks to achieve robust association. It combines the strengths of motion and appearance information, and also incorporates camera-motion compensation and a more accurate Kalman filter state vector.
COCO-Text~\cite{veit2016coco}, RCTW17~\cite{shi2017icdar2017}, ArT~\cite{chng2019icdar2019}, LSVT~\cite{sun2019chinese}, LSVTD~\cite{cheng2019you}, as the public datasets, are used in the training stage.
RandomHorizontalFlip, RandomRotate, ColorJitter, MotionBlur and GaussNoise were used for data augmentation.

\textbf{AI Lab, Du Xiaoman Financial.} The team selected TransDETR~\cite{wu2022end} as the baseline and employed the public datasets COCO-Text V2.0~\cite{veit2016coco} and SynthText~\cite{synthtext} as the pre-training data. To enhance the model's capacity to detect small texts, additional small texts were added to the SynthText images. Furthermore, the HRNet~\cite{wang2020deep} was employed as the new backbone, which demonstrated superiority in identifying faint text objects.
The team modify the original hyper-parameters of TransDETR to detect more texts from a single
frame. When loading the training data, the maximum number of text instance queries of the Transformer module is set to 400. 

\subsection{Top 3 Submissions in Task 2}

\textbf{Tencent-OCR team.}  
To enable end-to-end video text spotting, two methods, namely Parseq~\cite{bautista2022scene} and ABINet~\cite{fang2021read}, were utilized in the recognition stage. Both methods were trained on a dataset of 20 million samples, extracted from various open-source datasets, including ICDAR-2013~\cite{karatzas2013icdar}, ICDAR-2015~\cite{karatzas2015icdar}, COCO-Text~\cite{veit2016coco}, SynthText~\cite{synthtext}, among others.

During the end-to-end text spotting stage, different recognition methods are applied to predict all the detected boxes of a trajectory. The final text result corresponding to the trajectory is selected based on confidence and character length. Trajectories with low-quality text results, indicated by low scores or containing only one character, are removed.

\textbf{Guangzhou Shiyuan Technology team.} The text tracking task was addressed in a similar manner to Task 1. To recognize text, the PARSeq method was employed, which involves learning an ensemble of internal autoregressive (AR) language models with shared weights using Permutation Language Modeling. This approach unifies context-free non-AR and context-aware AR inference, along with iterative refinement using bidirectional context. The recognition model was trained using several extra public datasets, including COCO-Text~\cite{veit2016coco}, RCTW17~\cite{shi2017icdar2017}, ArT~\cite{chng2019icdar2019}, LSVT~\cite{sun2019chinese}, and LSVTD~\cite{cheng2019you}.

\textbf{China Mobile Communications Research Institute team.} The team used CoText~\cite{wu2022real} as the baseline and utilized ABINet~\cite{fang2021read} to enhance the recognition head. The Cotext model was trained using the ICDAR2015~\cite{karatzas2015icdar}, ICDAR2015Video~\cite{karatzas2015icdar}, and ICDAR2023 DSText datasets for text detection and tracking. The recognition part employed a pretrained ABINet model based on the MJSynth and SynthText~\cite{synthtext} datasets.

\section{Discussion}
\textbf{Text tracking task.} In this task,  most participants firstly employ the powerful backbone to enhance the performance, \eg{} HRNet, Res2Net and SENet.
With multiple backbones, TencentOCR team achieves the best score in three main metrics, \ie{} $\rm MOTA$, $\rm MOTP$, and ${\rm ID_{F1}}$/\%, as shown in Table.~\ref{table2}. 
For the text tracking, based on ByteTrack, the team designed four different metrics to compare the matching similarity between the current frame detection box and the existing text trajectory, i.e., box IoU, text content similarity, box size proximity and text geometric neighborhood relationship.
To further enhance the performance of result, most participants use the various data augmentations, \eg{} random motion blur, random rotation, random crop.
Besides, various public datasets, \eg{} COCO-Text~\cite{veit2016coco}, RCTW17~\cite{shi2017icdar2017}, ArT~\cite{chng2019icdar2019}, LSVT~\cite{sun2019chinese}, and LSVTD~\cite{cheng2019you} are used for joint training.

\textbf{End-to-End Video Text Spotting task.}
To enhance the end-to-end text spotting, most participants adopted the advanced recognition models, \ie{} Parseq~\cite{bautista2022scene} and ABINet~\cite{fang2021read}.
Large synthetic datasets (\eg{} SynthText~\cite{synthtext}), firstly are used to pretrain the model, and then further finetuned on the released training dataset (DSText).
With various data augmentation, large public datasets, powerful network backbones and
model ensembles, TencentOCR team achieves the best score in three main metrics, as shown in Table.~\ref{table3}.

Overall, while many participants implemented various improvement techniques such as using extra datasets and data augmentation, the majority of their results were unsatisfactory, with MOTA scores below $25\%$ and ${\rm ID_{F1}}$ scores below $70\%$. As a result, there is still a significant amount of room for improvement in this benchmark and many technical challenges to overcome.
It is worth mentioning that many of the top ranking methods utilize an ensemble of multiple models and large public datasets to enhance their performance. However, these pipelines tend to be complex and the corresponding inference speeds are slow. Simplifying the pipeline and accelerating inference are also important considerations for the video text spotting task.
Additionally, it is noteworthy that many of the submitted methods adopted different ideas and strategies, providing the community with new insights and potential solutions. We expect that more innovative approaches will be proposed following this competition.

\section{Conclusion}

Here, we present a new video text reading benchmark, which focuses on dense and small video text.
Compare with the previous datasets, the proposed dataset mainly includes two new challenges for dense and small video text spotting. 
High-proportioned small texts are a new challenge for the existing video text methods.
Meanwhile, we also organize the corresponding competition on Robust Reading Competition Website~\footnote{https://rrc.cvc.uab.es/?ch=22\&com=introduction}, where we received around 30 valid submissions from 24 teams.
These submissions will provide the community with new insights and potential solutions.
Overall, we believe and hope the benchmark, as one standard benchmark, develops and improve the video text tasks in the community.

\section{Potential Negative Societal Impacts and Solution}
Similar to BOVText~\cite{wu2021bilingual}, we blur the human faces in DSText with two steps. Firstly, detecting human faces in each frame with \textit{face recognition}\footnote{https://github.com/ageitgey/face\_recognition} - an easy-to-use face recognition open source project with complete development documents and application cases. 
Secondly, after obtaining the detection box, we blur the face with Gaussian Blur operation in OpenCV\footnote{https://www.tutorialspoint.com/opencv/opencv\_gaussian\_blur.htm}.

\section{Acknowledgements} This competition is supported by the National Natural Science Foundation (NSFC\#62225603).

\section{Competition Organizers}
The benchmark is mainly done by Weijia Wu and Yuzhong Zhao, while they are research interns at Kuaishou Technology.
The establishment of the benchmark is supported by the annotation team of Kuaishou Technology.
Prof. Xiang Bai at Huazhong University of Science and Technology, Prof. Dimosthenis Karatzas at the Universitat Autónoma de Barcelona, Prof. Umapada Pal at Indian Statistical Institute, and Asst Prof. Mike Shou at the National University of Singapore, as the four main supervisors, provide many valuable suggestions and comments, \eg{} annotation format suggestion from Prof. Xiang Bai, competition schedule plan from Prof. Dimosthenis Karatzas, submission plan and suggestions for the proposal from Prof.Umapada Pal, and statistical analysis from Asst Prof. Mike Shou.
Therefore, our team mainly includes eight people from seven institutions.

\bibliographystyle{IEEEtran}
\bibliography{egbib}

%




\end{document}

%% file: table/table1.tex
\def\x{{$\footnotesize \times$}}
\small  
\setlength{\tabcolsep}{2.0pt}
\begin{tabular}{l|c|c|c|c|c|c|p{0.6\columnwidth}}
    \whline
	Dataset &  Video & Frame & Text & Box Type & Text Area~(\# pixels) & Text Density & Supported Scenario~(Domain) \cr\shline \hline
	\hline
    YVT\cite{nguyen2014video}  & 30 & 13k & 16k & Upright& 8,664 & 1.15 & Cartoon, Outdoor(supermarket, shopping street, driving...) 
    \cr\cline{8-8} 
    ICD15 VT\cite{zhou2015icdar} & 51 & 27k & 144k & Oriented & 5,013 & 5.33 &  Driving, Supermarket, Shopping street...
    \cr\cline{8-8} 
    RoadText-1K\cite{reddy2020roadtext} & 1k & 300k & 1.2m & Upright & 2,141 & 0.75 & Driving
    \cr\cline{8-8} 
    LSVTD\cite{cheng2019you} & 100 & 66k & 569k & Oriented & 2,254 & 5.52 & Shopping mall, Supermarket, Hotel...
    \cr\cline{8-8} 
    BOVText~\cite{wu2021bilingual} & 2k & 1.7m & 8.8m & Oriented & 10,309 & 5.12 &    Cartoon,~Vlog, Travel, Game, Sport, News ...
    \\
    \hline
    DSText & 100 & 56k & 671k & Oriented & \textbf{1,984} & \textbf{23.5} &  Driving, Activity, Vlog, Street View~(indoor), Street View~(outdoor), Travel, News, Movie, Cooking\\
    \whline
     
\end{tabular}

%% file: table/task1.tex
\def\x{{$\footnotesize \times$}}
\scriptsize
\setlength{\tabcolsep}{7pt}
\begin{tabular}{c|c|ccc|ccc|p{0.4\columnwidth}}
    \whline
	User ID  & Rank &  $\rm MOTA$& $\rm MOTP$ & ${\rm ID_{F1}}$/\% & 	Mostly Matched &  Partially Matched & 	Mostly Lost &Affiliations \cr\shline \hline
	\hline
	TencentOCR
     & 1 & 	62.56\% & 79.88\% & 75.87\% & 8114	& 1800&	2663&	 Tencent. \\
     DA
     & 2 & 	50.52\% & 78.33\% & 	70.99\% & 7121	& 2405 &	3051&	 Guangzhou Shiyuan Electronic Technology Company Limited \\
     Tianyu Zhang
     & 3 & 		43.52\% & 78.15\% & 	62.27\% & 4980	& 2264 &	5333&	 AI Lab, Du Xiaoman Financial \\
     Liu Hongen
     & 4 & 		36.87\% & 79.24\% & 	48.99\% & 2123	& 3625 &	6829 &	 Tianjin Univeristy \\
     Yu Hao
     & 5 & 		31.01\% & 78.00\% & 		50.39\% & 2361	& 1767 &	8449 &	 - \\
     Hu Jijin
     & 6 & 		28.92\% & 78.46\% & 		43.96\% & 1385	& 1186 &	10006 &	 Beijing University of Posts \& Telecommunications \\
     	CccJ
     & 7 & 		27.55\% & 78.40\% & 		44.28\% & 1583	& 1103 &	9891 &	 CQUT \\
     MiniDragon
     & 8 & 		25.75\% & 	74.03\% & 		50.22\% & 3302	& 2806 &	6469 &	 - \\
     FanZhengDuo
     & 9 & 		23.41\% & 	75.54\% & 			49.66\% & 5216	& 3578 &	3783 &	 - \\
     zjb
     &10 & 		19.85\% & 		71.98\% & 			39.87\% & 2815	& 3354 &	6408 &	 - \\
     dunaichao
     &11 & 		19.84\% & 		73.82\% & 			31.18\% & 924	& 1765 &	9888 &	 China Mobile Communications Research Institute \\
     JiangQing
     &12 & 		13.83\% & 		75.75\% & 			58.41\% & 6924	& 2622 &	3031 &	 South China University of Technology; Shanghai AI Laboratory; KingSoft Office CV\&D Department \\
     Kebin Liu
     &13 & 			7.49\% & 		75.62\% & 		45.68\% & 5403		& 3835 &	3339 &	 Beijing University of Posts and Telecommunications \\
     TungLX &14 & 			0\% & 		0\% & 		0\% & 0		& 0 &	0 &	 - \\
     \whline
\end{tabular}

%% file: table/task2.tex
\def\x{{$\footnotesize \times$}}
\scriptsize
\setlength{\tabcolsep}{7pt}
\begin{tabular}{c|c|ccc|ccc|p{0.4\columnwidth}}
    \whline
	User ID  & Rank &  $\rm MOTA$& $\rm MOTP$ & ${\rm ID_{F1}}$/\% & 	Mostly Matched &  Partially Matched & 	Mostly Lost &Affiliations \cr\shline \hline
	\hline
	TencentOCR
     & 1 & 	22.44\% & 80.82\% & 56.45\% & 5062	& 1075&	6440&	 Tencent. \\
     DA
     & 2 & 		10.51\% & 78.97\% & 	53.45\% & 4629	& 1392 &	6556&	 Guangzhou Shiyuan Electronic Technology Company Limited \\
     dunaichao
     &3 & 		5.54\% & 		74.61\% & 			24.25\% & 528	& 946 &	11103 &	 China Mobile Communications Research Institute \\
 	cnn\_lin &4 & 		0\% & 		0\% & 			0\% & 0	& 0 &	0 &	 South China Agricultural University \\
 	Hu Jijin &5 & 		0\% & 		0\% & 			0\% & 0	& 0 &	0 &	 Beijing University of Posts and Telecommunications \\
 	XUE CHUHUI &6 & 		0\% & 		0\% & 			0\% & 0	& 0 &	0 &	 - \\
 	MiniDragon &7 & 		-25.09\% & 		74.95\% & 			26.38\% & 1388	& 1127 &	10062 &	 - \\
 	JiangQing
     &8 & 		-27.47\% & 		76.59\% & 			43.61\% & 4090	& 1471 &	7016 &	 South China University of Technology; Shanghai AI Laboratory; KingSoft Office CV\&D Department \\
     Tianyu Zhang
     & 9 & 		-28.58\% & 80.36\% & 		26.20\% & 1556	& 543 &	10478&	 AI Lab, Du Xiaoman Financial \\
     \whline
\end{tabular}

%% file: egbib.bbl
\begin{thebibliography}{10}
\providecommand{\url}[1]{#1}
\csname url@samestyle\endcsname
\providecommand{\newblock}{\relax}
\providecommand{\bibinfo}[2]{#2}
\providecommand{\BIBentrySTDinterwordspacing}{\spaceskip=0pt\relax}
\providecommand{\BIBentryALTinterwordstretchfactor}{4}
\providecommand{\BIBentryALTinterwordspacing}{\spaceskip=\fontdimen2\font plus
\BIBentryALTinterwordstretchfactor\fontdimen3\font minus
  \fontdimen4\font\relax}
\providecommand{\BIBforeignlanguage}[2]{{%
\expandafter\ifx\csname l@#1\endcsname\relax
\typeout{** WARNING: IEEEtran.bst: No hyphenation pattern has been}%
\typeout{** loaded for the language `#1'. Using the pattern for}%
\typeout{** the default language instead.}%
\else
\language=\csname l@#1\endcsname
\fi
#2}}
\providecommand{\BIBdecl}{\relax}
\BIBdecl

\bibitem{yin2016text}
X.-C. Yin, Z.-Y. Zuo, S.~Tian, and C.-L. Liu, ``Text detection, tracking and
  recognition in video: a comprehensive survey,'' \emph{{IEEE} Transactions on
  Image Processing}, vol.~25, no.~6, pp. 2752--2773, 2016.

\bibitem{srivastava2015unsupervised}
N.~Srivastava, E.~Mansimov, and R.~Salakhudinov, ``Unsupervised learning of
  video representations using lstms,'' in \emph{International Conference on
  Machine Learning}, 2015, pp. 843--852.

\bibitem{dong2021dual}
J.~Dong, X.~Li, C.~Xu, X.~Yang, G.~Yang, X.~Wang, and M.~Wang, ``Dual encoding
  for video retrieval by text,'' \emph{IEEE Transactions on Pattern Analysis
  and Machine Intelligence}, 2021.

\bibitem{anagnostopoulos2008license}
C.-N.~E. Anagnostopoulos, I.~E. Anagnostopoulos, I.~D. Psoroulas, V.~Loumos,
  and E.~Kayafas, ``License plate recognition from still images and video
  sequences: A survey,'' \emph{IEEE Transactions on intelligent transportation
  systems}, vol.~9, no.~3, pp. 377--391, 2008.

\bibitem{karatzas2015icdar}
D.~Karatzas, L.~Gomez-Bigorda, A.~Nicolaou, S.~Ghosh, A.~Bagdanov, M.~Iwamura,
  J.~Matas, L.~Neumann, V.~R. Chandrasekhar, S.~Lu \emph{et~al.}, ``Icdar 2015
  competition on robust reading,'' in \emph{{IEEE} International Conference on
  Document Analysis and Recognition}, 2015, pp. 1156--1160.

\bibitem{nguyen2014video}
P.~X. Nguyen, K.~Wang, and S.~Belongie, ``Video text detection and recognition:
  Dataset and benchmark,'' in \emph{{IEEE} winter conference on applications of
  computer vision}, 2014, pp. 776--783.

\bibitem{reddy2020roadtext}
S.~Reddy, M.~Mathew, L.~Gomez, M.~Rusinol, D.~Karatzas, and C.~Jawahar,
  ``Roadtext-1k: Text detection \& recognition dataset for driving videos,'' in
  \emph{{IEEE} International Conference on Robotics and Automation}, 2020, pp.
  11\,074--11\,080.

\bibitem{cheng2019you}
Z.~Cheng, J.~Lu, Y.~Niu, S.~Pu, F.~Wu, and S.~Zhou, ``You only recognize once:
  Towards fast video text spotting,'' in \emph{{ACM} International Conference
  on Multimedia}, 2019, pp. 855--863.

\bibitem{wu2021bilingual}
W.~Wu, D.~Zhang, Y.~Cai, S.~Wang, J.~Li, Z.~Li, Y.~Tang, and H.~Zhou, ``A
  bilingual, openworld video text dataset and end-to-end video text spotter
  with transformer,'' in \emph{Thirty-fifth Conference on Neural Information
  Processing Systems Datasets and Benchmarks Track (Round 2)}, 2021.

\bibitem{zhou2015icdar}
X.~Zhou, S.~Zhou, C.~Yao, Z.~Cao, and Q.~Yin, ``Icdar 2015 text reading in the
  wild competition,'' \emph{arXiv preprint arXiv:1506.03184}, 2015.

\bibitem{dendorfer2019cvpr19}
P.~Dendorfer, H.~Rezatofighi, A.~Milan, J.~Shi, D.~Cremers, I.~Reid, S.~Roth,
  K.~Schindler, and L.~Leal-Taixe, ``Cvpr19 tracking and detection challenge:
  How crowded can it get?'' \emph{arXiv preprint arXiv:1906.04567}, 2019.

\bibitem{karatzas2013icdar}
D.~Karatzas, F.~Shafait, S.~Uchida, M.~Iwamura, L.~G. i~Bigorda, S.~R. Mestre,
  J.~Mas, D.~F. Mota, J.~A. Almazan, and L.~P. De~Las~Heras, ``Icdar 2013
  robust reading competition,'' in \emph{2013 12th international conference on
  document analysis and recognition}.\hskip 1em plus 0.5em minus 0.4em\relax
  IEEE, 2013, pp. 1484--1493.

\bibitem{li2009learning}
Y.~Li, C.~Huang, and R.~Nevatia, ``Learning to associate: Hybridboosted
  multi-target tracker for crowded scene,'' in \emph{2009 IEEE conference on
  computer vision and pattern recognition}.\hskip 1em plus 0.5em minus
  0.4em\relax IEEE, 2009, pp. 2953--2960.

\bibitem{ristani2016performance}
E.~Ristani, F.~Solera, R.~Zou, R.~Cucchiara, and C.~Tomasi, ``Performance
  measures and a data set for multi-target, multi-camera tracking,'' in
  \emph{Workshops of European conference on computer vision}, 2016, pp. 17--35.

\bibitem{wu2022end}
W.~Wu, D.~Zhang, Y.~Fu, C.~Shen, H.~Zhou, Y.~Cai, and P.~Luo, ``End-to-end
  video text spotting with transformer,'' \emph{arXiv preprint
  arXiv:2203.10539}, 2022.

\bibitem{cai2018cascade}
Z.~Cai and N.~Vasconcelos, ``Cascade r-cnn: Delving into high quality object
  detection,'' in \emph{Proceedings of the IEEE conference on computer vision
  and pattern recognition}, 2018, pp. 6154--6162.

\bibitem{wang2020deep}
J.~Wang, K.~Sun, T.~Cheng, B.~Jiang, C.~Deng, Y.~Zhao, D.~Liu, Y.~Mu, M.~Tan,
  X.~Wang \emph{et~al.}, ``Deep high-resolution representation learning for
  visual recognition,'' \emph{IEEE transactions on pattern analysis and machine
  intelligence}, vol.~43, no.~10, pp. 3349--3364, 2020.

\bibitem{wang2022internimage}
W.~Wang, J.~Dai, Z.~Chen, Z.~Huang, Z.~Li, X.~Zhu, X.~Hu, T.~Lu, L.~Lu, H.~Li
  \emph{et~al.}, ``Internimage: Exploring large-scale vision foundation models
  with deformable convolutions,'' \emph{arXiv preprint arXiv:2211.05778}, 2022.

\bibitem{zhang2022bytetrack}
Y.~Zhang, P.~Sun, Y.~Jiang, D.~Yu, F.~Weng, Z.~Yuan, P.~Luo, W.~Liu, and
  X.~Wang, ``Bytetrack: Multi-object tracking by associating every detection
  box,'' in \emph{Computer Vision--ECCV 2022: 17th European Conference, Tel
  Aviv, Israel, October 23--27, 2022, Proceedings, Part XXII}.\hskip 1em plus
  0.5em minus 0.4em\relax Springer, 2022, pp. 1--21.

\bibitem{synthtext}
A.~Gupta, A.~Vedaldi, and A.~Zisserman, ``Synthetic data for text localisation
  in natural images,'' in \emph{{IEEE} Conference on Computer Vision and
  Pattern Recognition}, 2016, pp. 2315--2324.

\bibitem{he2017mask}
K.~He, G.~Gkioxari, P.~Doll{\'a}r, and R.~Girshick, ``Mask r-cnn,'' in
  \emph{Proceedings of the IEEE international conference on computer vision},
  2017, pp. 2961--2969.

\bibitem{liao2020real}
M.~Liao, Z.~Wan, C.~Yao, K.~Chen, and X.~Bai, ``Real-time scene text detection
  with differentiable binarization,'' in \emph{Proceedings of the AAAI
  conference on artificial intelligence}, vol.~34, no.~07, 2020, pp.
  11\,474--11\,481.

\bibitem{gao2021video}
Y.~Gao, X.~Li, J.~Zhang, Y.~Zhou, D.~Jin, J.~Wang, S.~Zhu, and X.~Bai, ``Video
  text tracking with a spatio-temporal complementary model,'' \emph{IEEE
  Transactions on Image Processing}, vol.~30, pp. 9321--9331, 2021.

\bibitem{aharon2022bot}
N.~Aharon, R.~Orfaig, and B.-Z. Bobrovsky, ``Bot-sort: Robust associations
  multi-pedestrian tracking,'' \emph{arXiv preprint arXiv:2206.14651}, 2022.

\bibitem{veit2016coco}
A.~Veit, T.~Matera, L.~Neumann, J.~Matas, and S.~Belongie, ``Coco-text: Dataset
  and benchmark for text detection and recognition in natural images,''
  \emph{arXiv preprint arXiv:1601.07140}, 2016.

\bibitem{shi2017icdar2017}
B.~Shi, C.~Yao, M.~Liao, M.~Yang, P.~Xu, L.~Cui, S.~Belongie, S.~Lu, and
  X.~Bai, ``Icdar2017 competition on reading chinese text in the wild
  (rctw-17),'' in \emph{2017 14th iapr international conference on document
  analysis and recognition (ICDAR)}, vol.~1.\hskip 1em plus 0.5em minus
  0.4em\relax IEEE, 2017, pp. 1429--1434.

\bibitem{chng2019icdar2019}
C.~K. Chng, Y.~Liu, Y.~Sun, C.~C. Ng, C.~Luo, Z.~Ni, C.~Fang, S.~Zhang, J.~Han,
  E.~Ding \emph{et~al.}, ``Icdar2019 robust reading challenge on
  arbitrary-shaped text-rrc-art,'' in \emph{2019 International Conference on
  Document Analysis and Recognition (ICDAR)}.\hskip 1em plus 0.5em minus
  0.4em\relax IEEE, 2019, pp. 1571--1576.

\bibitem{sun2019chinese}
Y.~Sun, J.~Liu, W.~Liu, J.~Han, E.~Ding, and J.~Liu, ``Chinese street view
  text: Large-scale chinese text reading with partially supervised learning,''
  in \emph{Proceedings of the IEEE/CVF International Conference on Computer
  Vision}, 2019, pp. 9086--9095.

\bibitem{bautista2022scene}
D.~Bautista and R.~Atienza, ``Scene text recognition with permuted
  autoregressive sequence models,'' in \emph{Computer Vision--ECCV 2022: 17th
  European Conference, Tel Aviv, Israel, October 23--27, 2022, Proceedings,
  Part XXVIII}.\hskip 1em plus 0.5em minus 0.4em\relax Springer, 2022, pp.
  178--196.

\bibitem{fang2021read}
S.~Fang, H.~Xie, Y.~Wang, Z.~Mao, and Y.~Zhang, ``Read like humans: Autonomous,
  bidirectional and iterative language modeling for scene text recognition,''
  in \emph{Proceedings of the IEEE/CVF Conference on Computer Vision and
  Pattern Recognition}, 2021, pp. 7098--7107.

\bibitem{wu2022real}
W.~Wu, Z.~Li, J.~Li, C.~Shen, H.~Zhou, S.~Li, Z.~Wang, and P.~Luo, ``Real-time
  end-to-end video text spotter with contrastive representation learning,''
  \emph{arXiv preprint arXiv:2207.08417}, 2022.

\end{thebibliography}
